%% file: main.tex
\documentclass[lettersize,journal]{IEEEtran}
\usepackage{amsmath,amsfonts}
\usepackage{algorithmic}
\usepackage{array}
\usepackage[caption=false,font=normalsize,labelfont=sf,textfont=sf]{subfig}
\usepackage{textcomp}
\usepackage{stfloats}
\usepackage{url}
\usepackage{verbatim}
\usepackage{graphicx}
\def\BibTeX{{\rm B\kern-.05em{\sc i\kern-.025em b}\kern-.08em
    T\kern-.1667em\lower.7ex\hbox{E}\kern-.125emX}}
\usepackage{balance}

\usepackage{fancyhdr}

\usepackage[abbreviations]{glossaries-extra}
\glsdisablehyper

\glssetcategoryattribute{acronym}{indexonlyfirst}{true}

\newabbreviation{cnn}{CNN}{Convolutional Neural Network}
\newabbreviation{vlm}{VLM}{vision-language model}
\newabbreviation{vlms}{VLMs}{vision-language models}
\newabbreviation{clip}{CLIP}{Contrastive Language–Image Pre-training}
\newabbreviation{sar}{SAR}{Synthetic-aperture radar}

\begin{document}
\title{Exploring the Capability of Text-to-Image Diffusion Models with Structural Edge Guidance\\for Multi-Spectral Satellite Image Inpainting}
\author{Mikolaj Czerkawski, Christos Tachtatzis\\Department of Electronic and Electrical Engineering\\University of Strathclyde, Glasgow, UK}

\fancyhf{}
\renewcommand{\headrulewidth}{0pt}
\fancyfoot[c]{}
\fancypagestyle{FirstPage}{
\lfoot{DOI: \url{10.1109/LGRS.2024.3370212} \copyright2024 IEEE}
\rfoot{IEEE Geoscience and Remote Sensing Letters}
}

\maketitle
\thispagestyle{FirstPage}

\begin{abstract}
    The paper investigates the utility of text-to-image inpainting models for satellite image data. Two technical challenges of injecting structural guiding signals into the generative process as well as translating the inpainted RGB pixels to a wider set of MSI bands are addressed by introducing a novel inpainting framework based on StableDiffusion and ControlNet as well as a novel method for RGB-to-MSI translation. The results on a wider set of data suggest that the inpainting synthesized via StableDiffusion suffers from undesired artefacts and that a simple alternative of self-supervised internal inpainting achieves a higher quality of synthesis. Source code: \url{}
\end{abstract}

\begin{IEEEkeywords}
image inpainting, image completion, generative models
\end{IEEEkeywords}

\section{Introduction}
\label{sec:intro}

    \IEEEPARstart{O}{ftentimes}, it is desired inpaint regions of optical satellite images, due to common issues such as a sensor failure or natural conditions like cloud cover. This challenge has been widely addressed in the literature~\cite{Singh2018,Meraner2020,Ebel2021,Xu2022,Sebastianelli2022}, however, the use of general pre-trained text-to-image diffusion models to solve this problem has not yet been explored.

    It is not currently clear how beneficial these models, such as StableDiffusion~\cite{Rombach2022}, could be for this task. Trained on a diverse set of data (containing both text and images) and optimized on the challenging task of text-to-image synthesis, they could be a source of meaningful inductive bias. Furthermore, denoising diffusion models bring some advantages of their own, such as the ability to trade-off compute cost and synthesis quality at inference time~\cite{song2023consistency}. Furthermore, StableDiffusion~\cite{Rombach2022} (which can be used for inpainting) is also compatible with ControlNet~\cite{Zhang2023}, permitting the injection of a structural guidance image into the synthesis process, allowing to condition the reconstruction on the historical satellite data.

\begin{figure*}
        \centering
        \includegraphics[width=\textwidth]{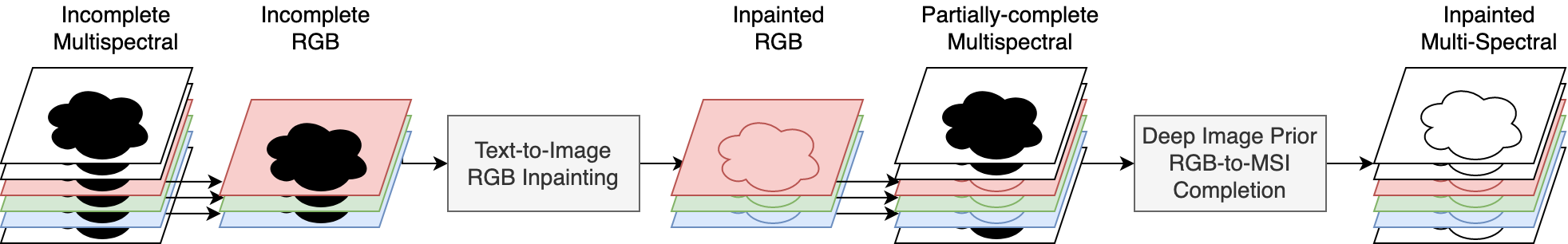}
        \caption{Complete pipeline for multi-spectral satellite image inpainting. The process is built on a sequence of two steps, where a pre-trained diffusion model is first applied to RGB data for inpainting (text-to-image RGB inapinting), and then a Deep Image Prior~\cite{Ulyanov2020} approach is used to transfer that the reconstruction beyond the non-RGB channels (Deep Image Prior RGB-to-MSI Completion). This allows for any RGB-based pre-trained model, such as StableDiffusion~\cite{Rombach2022} to be incorporated into a pipeline capable of mutli-spectral satellite image inpainting.}
        \label{fig:dip-diagram}
    \end{figure*}

    In this manuscript, two solutions are proposed and tested using off-the-shelf open-source text-to-image models to inpaint large portions of multi-spectral 13-band Sentinel-2 images, by employing a two-stage approach consisting of (1) RGB-based inpainting and (2) a novel approach to RGB-to-MSI zero-shot translation, as illustrated in Figure~\ref{fig:dip-diagram}. A set of experiments with various hyperparameter settings is provided to gain an understanding of the influence these have on the proposed framework. This is followed by an evaluation of the proposed inpainting methods on multi-spectral satellite image data.

\section{Related Work}
\label{sec:rel_work}

    This work considers methods that can be freely applied to multi-spectral image representations and adapt to their content and channel count (assuming their representation includes the RGB channels). Existing pre-trained models, such as the one based on convolutional neural processes for inpainting~\cite{Pondaven2022}, do not provide such a level of freedom and hence remain out of the scope of this work.

    \textbf{Denoising Diffusion Models.} The approach to use denoising diffusion processes for image generation has become popular recently~\cite{Sohl-Dickstein2015,Ho2020,Song2020,Nichol2021,Dhariwal2021}. While more expensive at inference, diffusion models have been shown to beat the state of the art in terms of image quality~\cite{Dhariwal2021}. Other advantages include control over the trade-off between quality and inference time, as well as ways of injecting many types of guidance to condition the synthesis~\cite{Dhariwal2021}. Text prompts are a key method for conditioning, giving rise to text-to-image models such as DALLE-2~\cite{Ramesh2022}, Imagen~\cite{Saharia2022}, or StableDiffusion~\cite{Rombach2022}. The latter, StableDiffusion, is the first open-source implementation of such a model with free and open access to model weights, which this work benefits from.

    \textbf{Single Image Synthesis for Satellite Images.} Solutions relying on learning based on the input sample only can often be applied to any type of data modality and shape. For the task of satellite image inpainting, several works have explored this direction~\cite{Ebel2021,Czerkawski2022} by relying on priors captured by the a convolutional neural network topology, along the lines of the seminal work on Deep Image Prior~\cite{Ulyanov2020}. However, none of the past approaches has explored the potential of combining single image synthesis with the general-purpose text-to-image diffusion models.

\section{Method}
\label{sec:method}

    \subsection{Components}

    \textbf{StableDiffusion} is a variant of latent diffusion~\cite{Rombach2022} focused on the text conditioning modality. Latent diffusion is a special type of image diffusion aimed at high-resolution data, which employs an autoencoder to compress image input into a latent space so that the denoising diffusion process is performed in a more compact domain than high-resolution images. StableDiffusion refers not only to the method (which is technically the same as latent diffusion) but also to the model weights released by the StabilityAI organisation. So far, several models have been released, with different levels of performance, but also, as in the case of StableDiffusion-2, a different set of potential inputs; for example, depth image condition. In this work, the StableDiffusion1.5 checkpoint is used with the model variant trained to perform inpainting based on an additional inpainting mask provided in input.

    \textbf{Diffusion for Inpainting.} Several approaches for employing denoising diffusion for the task of image inpainting have already been considered. Some prominent examples include Palette~\cite{Saharia2022}, where the network performing denoising diffusion in pixel space is also provided with an additional channel corresponding to the inpainting mask, effectively serving as an extra condition present in the input. An alternative approach of RePaint~\cite{Lugmayr2022} uses a pre-trained denoising diffusion model for inpainting with a different type of sampling scenario, where the known regions of the input (passed through the forward process) are mixed with the diffused signal representation. Finally, StableDiffusion also provides inpainting-oriented variants, where the underlying latent space model accepts an extra channel for the downsampled mask besides the usual 4-channel latent representation~\cite{Rombach2022}. As mentioned earlier, the same approach was used for the inpainting model of StableDiffusion, the inpainting core model for this work.

    \textbf{ControlNet.} Another important component of the presented approach is ControlNet, which enables to inject additional spatial guidance into the inpainting process. ControlNet has been introduced~\cite{Zhang2023}  an extension to StableDiffusion with the aim of incorporating more image-based conditions into the synthesis process. More specifically, the method uses a separate encoding network to encode latents based on a preselected condition type and mixes the encoded representations with the internal representations in the core StableDiffusion model. To ensure preservation of the features learned by the core StableDiffusion model, a zero-convolution technique has been used to nullify the effect of the added network at the beginning of the fine-tuning process. In the seminal ControlNet paper~\cite{Zhang2023}, several choices for the conditioning signal type are proposed, including Canny edge, Hough lines, user sketches, human pose, or HED boundary detections~\cite{Xie2017}.

    \subsection{Edge-Guided Inpainting for RGB.} 
\begin{figure*}
        \centering
        \includegraphics[width=0.8\textwidth]{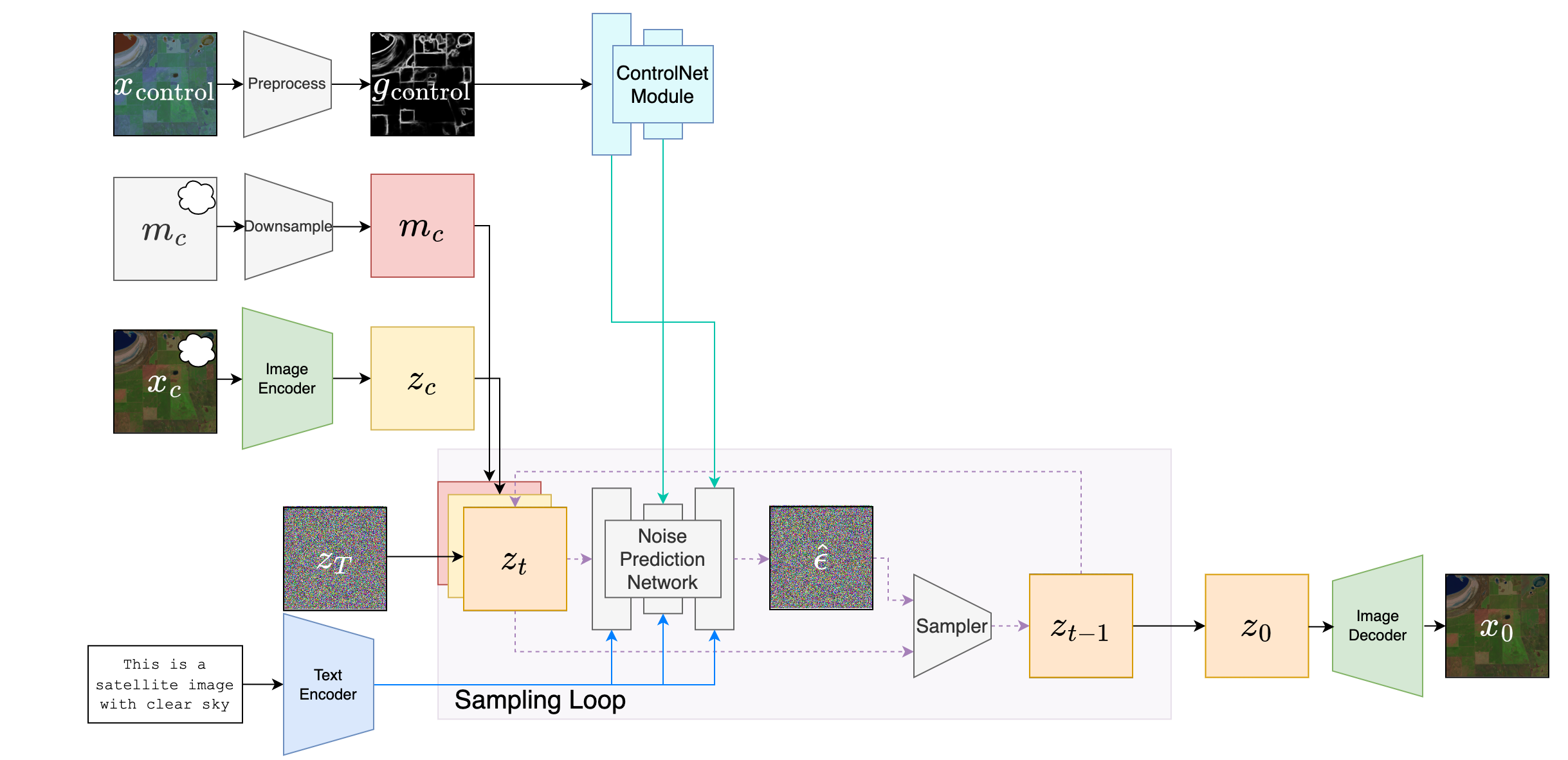}
        \caption{The \textbf{Edge-Guided Inpainting} diffusion pipeline used for this work employs a ControlNet approach~\cite{Zhang2023}, with an inpainting StableDiffusion backbone.}
        \label{fig:main-diagram}
    \end{figure*}

This work combines the inpainting capability of the StableDiffusion model with the edge-guidance capability of a HED boundary version of ControlNet, named \textbf{Edge-Guided Inpainting}. As shown in Figure~\ref{fig:main-diagram}, this is achieved by combining a pre-trained ControlNet model with the StableDiffusion inpainting backbone. This allows synthesis of the inpainted area, based on the provided mask input, control image and an optional text-prompt. This work focuses primarily on the use case with a historical edge-guidance, and a simple prompt of "a cloud-free satellite image". However, the proposed framework provides a large degree of flexibility between choosing different text-prompts (including negative prompts) to guide the output as well as different sources of structural information other than historical image edges.

    \subsection{RGB-to-MSI transfer with Deep Image Prior}

    The problem of generating multi-spectral images from an RGB sample has been explored in earlier works such as~\cite{Nguyen2014,Wu2017,Han2018,Zeng2021}, however, these solutions generally require training a specialised network for every image modality and are not well adjusted for this case, where part of the multi-spectral image is known. To enable a flexible, representation-agnostic approach, a Deep Image Prior~\cite{Ulyanov2020} technique is used, where a randomly initialized convolutional network is optimized to produce the known pixels in the output, similar to previous work on satellite image inpainting~\cite{Czerkawski2022}. A SkipNetwork with the same architecture as in~\cite{Czerkawski2022} is optimized with MSE loss backpropagated from the known region for 4,000 gradient steps at a learning rate of 0.02 (on a standard grade GPU, it takes about 2 minutes to process a single sample). The known region contains the pixels of all bands that are not missing as well as a complete inpainted image in the RGB bands, arranged in the same fashion as the input representation.

    \begin{table*}[]
            \centering
            \caption{Parameter test results for the text-based models.}
            \begin{tabular}{|c c|c c c c c c c c|}
                \hline
                & & \multicolumn{4}{c}{StableDiffusion Inpainting} & \multicolumn{4}{c|}{Edge-Guided Inpainting} \\
                & & \multicolumn{2}{c}{SSIM ($\uparrow$)} & \multicolumn{2}{c}{RMSE ($\downarrow$)} & \multicolumn{2}{c}{SSIM ($\uparrow$)} & \multicolumn{2}{c|}{RMSE ($\downarrow$)}\\
                & & Whole & Mask & Whole & Mask & Whole & Mask & Whole & Mask \\
                \hline
                Mask Content & Blank & 0.70 & 0.54 & 0.11 & 0.15 & 0.71 & 0.58 & 0.10 & 0.14\\
                & \textbf{Historical} & 0.78 & 0.67 & 0.10 & 0.13 & 0.77 & 0.67 & 0.10 & 0.13\\
                \hline
                & 0.0 & 0.78 & 0.67 & 0.10 & 0.13 & 0.77 & 0.67 & 0.10 & 0.13\\
                Text-Guidance Scale & \textbf{1.0} & 0.78 & 0.67 & 0.10 & 0.13 & 0.77 & 0.67 & 0.10 & 0.13\\
                & 7.5 & 0.77 & 0.66 & 0.11 & 0.14 & 0.78 & 0.68 & 0.09 & 0.12 \\
                \hline
                & \textbf{20} & 0.78 & 0.67 & 0.10 & 0.13 & 0.77 & 0.67 & 0.10 & 0.13\\
                Sampling Steps & 50 & 0.77 & 0.66 & 0.10 & 0.13 & 0.77 & 0.66 & 0.10 & 0.13 \\
                & 100 & 0.77 & 0.66 & 0.10 & 0.13 & 0.76 & 0.66 & 0.10 & 0.13 \\
                \hline
                & 0.1 & \multicolumn{4}{c}{NA} & 0.78 & 0.67 & 0.10 & 0.13\\
                Edge-Guidance Scale & 0.5 & \multicolumn{4}{c}{NA} & 0.79 & 0.69 & 0.09 & 0.12 \\
                & \textbf{1.0} & \multicolumn{4}{c}{NA} & 0.77 & 0.67 & 0.10 & 0.13 \\
                \hline
            \end{tabular}
            \label{tab:param_test}
        \end{table*}

\section{Results}
\label{sec:results} 
    The test dataset used for this study is extracted from SEN12MS-CR-TS~\cite{sen12mscrts} containing 888 cloud-free Sentinel-2 test samples, each paired with another historical cloud-free Sentinel-2 sample from the exact same location, which is used as the historical structure guide signal. The raw data is subject to the same preprocessing as in the related dataset~\cite{Meraner2020}, followed by a clipping operation to constrain the samples to [0,1] range expected by StableDiffusion. It was also ensured that the mean value of the samples is not higher than 0.9 before the clipping operation to exclude saturated images from analysis.

    The two tested text-based models include the standard StableDiffusion inpainting approach and the proposed Edge-Guided Inpainting approach. The results first demonstrate the effects of key parameters used for the text-based generative diffusion models (Table~\ref{tab:param_test}) and then report on the achieved performance of the 2-stage pipeline on the multi-spectral ground truth (Table~\ref{tab:main_results}) and the 1-stage pipelines on the RGB ground truth (Table~\ref{tab:main_rgb_results}).

    \subsection{Parameter Tests}

        The following parameters of the text-to-image model are tested: 1) the content of the masked region 2) the text-guidance scale 3) the number of sampling steps (this affects the inference time, for example, 20 sampling steps take about 5 seconds on a single standard grade GPU) 4) the edge-guidance scale (which only applies to the edge-guided inpainting approach). The results are shown in Table~\ref{tab:param_test}. For each tested parameter, the remaining parameters have the values highlighted in bold.

\begin{figure*}
            \centering
            \includegraphics[width=\textwidth]{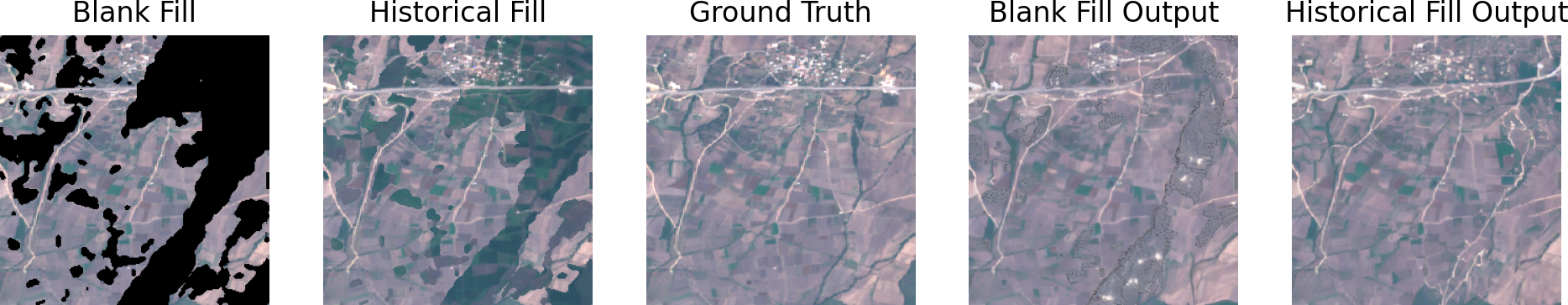}
            \caption{Comparison of the two methods of filling the masked region in the input to the diffusion models. Furthermore, output achieved with the StableDiffusion Inpainting scheme is shown for reference as a result of using each method.}
            \label{fig:mask_type}
        \end{figure*}

        Since this study is focused on the utility of historical optical data, explored is an approach to fill the missing regions with the values extracted from the historical sample instead of leaving them blank, in order to inject structure information into the network input. An example of these two input variants is shown in Figure~\ref{fig:mask_type}. 
        
        It is found that the historical input injection is beneficial and can improve the output structure, as indicated by the increased SSIM in Table~\ref{tab:param_test}. This effect occurs standard StableDiffusion inpainting (inpainted SSIM goes from 0.54 to 0.67) and Edge-Guided Inpainting (inpainted SSIM goes from 0.58 to 0.67). Without using this technique Edge-Guided inpainting performs better at reconstructing the missing region, owing to the additional guidance from the historical sample using ControlNet. However, the technique of historical input filling puts both standard and ControlNet inpainting variants on par.

        The classifier-free guidance scale for the text prompt, as defined in~\cite{Dhariwal2021}, controls the influence of the text input. and has been tested at three levels, ranging from 0.0 (no effective text guidance) through 1.0 to 7.5 (StableDiffusion default). Since additional information is supplied in the form of historical visual data (either via ControlNet or input historical filling), the differences are small. The text appears to be less useful for the standard StableDiffusion inpainting, with the performance slightly higher for low levels of text-guidance scale, while for the Edge-Guided Inpainting model it is the opposite.
 
        The UniPC sampler~\cite{Zhao2023} is used for all diffusion sampling and the main tested factor is the number of sampling steps, with tested values of 20 steps (reported in~\cite{Zhao2023} to yield good quality), 50 steps and 100 steps, verifying that the increased number of steps is not found to be beneficial for this task.

        The weight applied to the ControlNet features~\cite{Zhang2023} before they are added to the core network features is tested with the default value of 1.0, along with 0.5 and 0.1 values that explore a more subtle conditioning scheme. It is found that from the tested values, 0.5 achieves the highest output quality and is used for the subsequent experimentation.

        \begin{table}[h]
                \centering
                \caption{Inpainting Results computed \textbf{for all 13 channels} of the multispectral images in the test dataset.}
                \begin{tabular}{|l|cc|cc|}
                    \hline
                    Method & \multicolumn{2}{c|}{SSIM ($\uparrow$)} & \multicolumn{2}{c|}{RMSE ($\downarrow$)}\\
                    & Whole & Mask & Whole & Mask\\
                    \hline   
                    SD-Inpainting & 0.78 & 0.65 & 0.16 & 0.21 \\
                    Edge-Guided Inpainting& 0.62 & 0.48 & 0.37 & 0.48 \\
                    \hline
                    Direct-DIP & 0.64 & 0.45 & 0.38 & 0.53 \\
                    Direct-DIP w/ Historical & 0.85 & 0.74 & 0.14 & 0.19 \\
                    \hline
                    Ideal-RGB Channel Fill & 0.89 & 0.82 & 0.12 & 0.16\\
                    \hline
                \end{tabular}
                \label{tab:main_results}
            \end{table}
    
            \begin{table}[h]
                \centering
                \caption{Inpainting Results computed \textbf{only for the RGB channels} of the multispectral images in the test dataset.}
                \begin{tabular}{|l|cc|cc|}
                    \hline
                    Method & \multicolumn{2}{c|}{SSIM ($\uparrow$)} & \multicolumn{2}{c|}{RMSE ($\downarrow$)}\\
                    & Whole & Mask & Whole & Mask\\
                    \hline
                    SD-Inpainting & 0.78 & 0.67 & 0.10 & 0.13 \\
                    Edge-Guided Inpainting & 0.79 & 0.69 & 0.09 & 0.12 \\
                    \hline
                    Direct-DIP & 0.72 & 0.58 & 0.23 & 0.31 \\
                    Direct-DIP w/ Historical & 0.88 & 0.79 & 0.08 & 0.11 \\
                    \hline
                \end{tabular}
                \label{tab:main_rgb_results}
            \end{table}   

    \subsection{Multi-Spectral Inpainting Evaluation}

        The two proposed two-stage approaches based on text-to-image inpainting are compared against two single-stage methods applying direct inpainting with Deep Image Prior (similar to~\cite{Czerkawski2022}), with the performance on multi-spectral images listed in Table~\ref{tab:main_results}. In the last row, the RGB-to-MSI method is also tested (Ideal-RGB) with exact knowledge of RGB channels to approximate the maximum potential performance achievable with knowledge of the RGB channels, with 0.82 mask SSIM achieved. This demonstrates the value of the proposed RGB-to-MSI translation technique and its dependence on accurate reference inpainting data in the RGB channels. In practical terms, the best method is the Direct-DIP approach supplemented with historical data, with a SSIM of 0.74, followed by the text-to-image methods scoring 0.65 and 0.48, respectively.

        The advantage of the DIP-based benchmark is also observed for the RGB channels alone, as shown in Table~\ref{tab:main_rgb_results}.

        \begin{figure*}
            \centering
            \includegraphics[width=\textwidth]{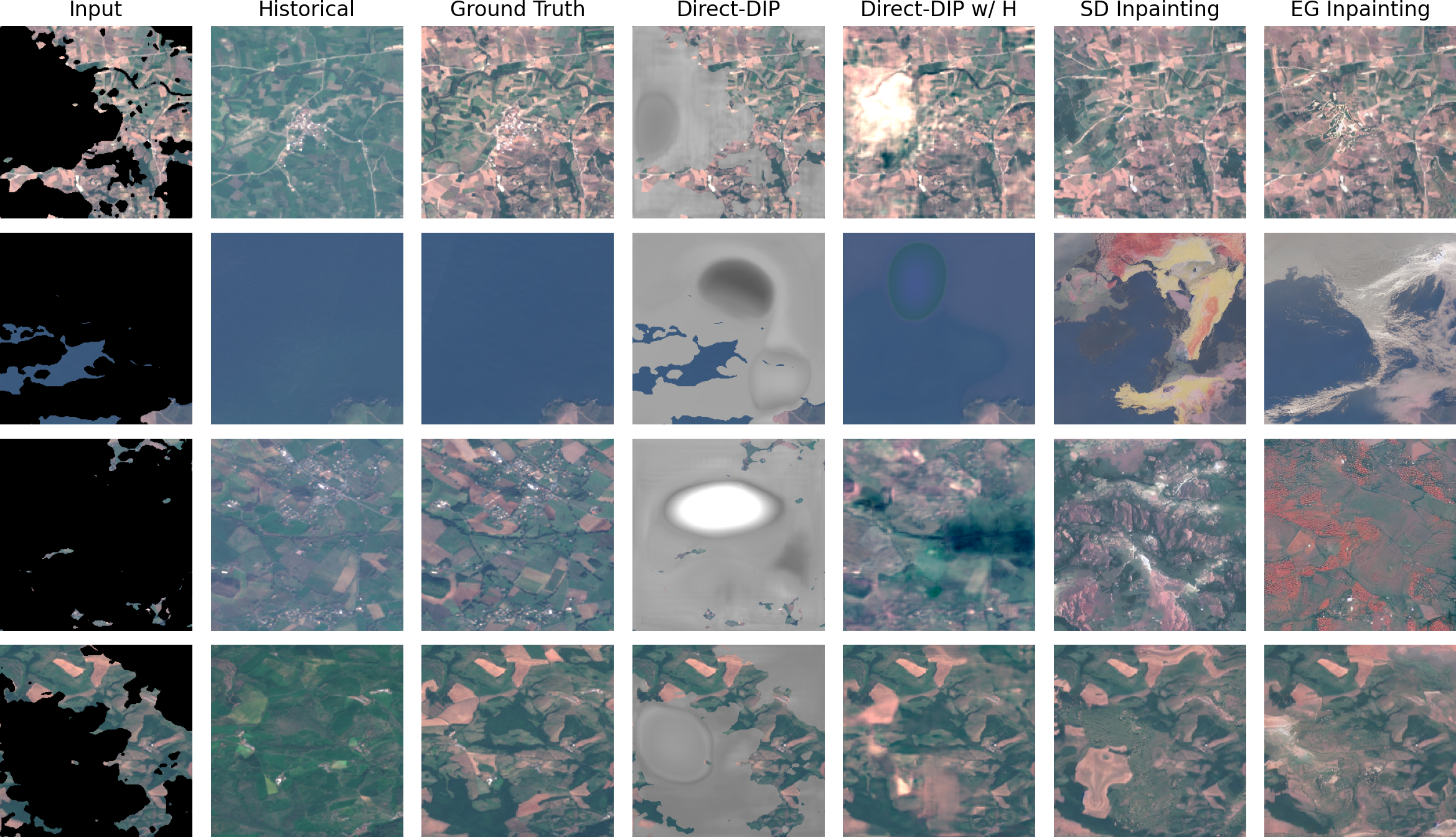}
            \caption{RGB visualization of 4 random samples drawn from the test dataset and the corresponding output from each method. It is shown that the Direct-DIP struggles to perform good quality inpainting with no extra source of information, producing visually incoherent output. The text-based models appear to produce visually coherent, yet inaccurate inpaintings, despite the efforts to inject correct structural information into the process.}
            \label{fig:rgb_comp}
        \end{figure*}
        
        The visual examples shown in Figure~\ref{fig:rgb_comp} illustrate the behavior of each framework, showing that for large inpainting masks, the text-to-image methods (last two rows) tend to introduce a lot of new objects into the scene, despite the structural guidance. For smaller masks, they appear to generate more convincing generations than DIP-based approaches.
    
\section{Conclusions}

    Even with structure-oriented adjustments, the general-purpose text-to-image diffusion models may not be immediately performant for the task of multi-spectral satellite image inpainting. Their synthetic capability is prone to generate unnatural artefacts in the output. Even a simple baseline of internal inpainting with historical data yields higher performance. This effect could be attributed to several factors, including the data models have been trained on (in the case of StableDiffusion~\cite{Rombach2022}, it was LAION-5B~\cite{Schuhmann2022laionb}) or the use of an autoencoder trained with losses focused on perceptual similarity~\cite{Zhang_2018_CVPR}. Despite the presence of Earth observation images in the training dataset of LAION-5B, they occupy a relatively small portion of the training domain and are often reprocessed in unpredictable ways (this is an effect of the dataset being scraped from the web)~\cite{LAION_EO}.

    Despite this limitation, the application of text-based models could inspire further work on the topic. Text offers a convenient conditioning mechanism, which could enable a variety of different applications, including a more controlled restoration process, or data augmentation. This work also introduced a method for elevating inpainted images performed in RGB (which can generally be expected to be a limitation of standard generative models) to the multi-spectral domain. In a circumstance, where the synthetic capability of the general-purpose generative models improves for the domain of satellite images, this presents a practical approach for transferring that capability to other sensing bands.

\input{sources.bbl}

\end{document}